\pdfoutput=1

\documentclass[11pt]{article}

\usepackage{acl}

\usepackage{times}
\usepackage{latexsym}

\usepackage[T1]{fontenc}

\usepackage[utf8]{inputenc}

\usepackage{microtype}

\usepackage{inconsolata}
\usepackage{import}
\usepackage{microtype}
\usepackage{layout}
\usepackage{tabularx, makecell}
\usepackage{booktabs}
\usepackage{mathrsfs}
\usepackage{amssymb} 
\usepackage{url}
\usepackage{graphicx}
\usepackage{xspace,paralist}
\usepackage{times,latexsym}
\usepackage{amsmath}
\usepackage{appendix}
\usepackage{comment} 
\usepackage{enumitem}
\usepackage{makecell}
\usepackage{multirow}
\usepackage{xcolor}
\usepackage{graphicx}
\usepackage{cleveref}
\usepackage{tcolorbox}
\usepackage{adjustbox}
\usepackage{relsize}
\usepackage{todonotes}

\usepackage[frozencache,cachedir=minted-cache]{minted} 
\usepackage{xcolor}

\setminted{
    frame=single,
    fontsize=\footnotesize,
    breaklines
}

\newcommand{\tabref}[2][]{Table#1~\ref{#2}\xspace}
\newcommand{\figref}[1]{Figure~\ref{#1}\xspace}
\newcommand{\secref}[1]{Section~\ref{#1}\xspace}
\newcommand{\appref}[1]{Appendix~\ref{#1}\xspace}

\newcommand{\errortype}[1]{\textit{#1}\xspace}
\newcommand{\ftype}{\errortype{Type1}}
\newcommand{\stype}{\errortype{Type2}}
\newcommand{\ttype}{\errortype{Type3}}

\newcommand{\model}[1]{\text{#1}\xspace}

\newcommand{\chatgpt}{\model{ChatGPT}}

\newcommand{\gptfouro}{\model{GPT-4o}}

\newcommand{\davinci}{\model{Davinci-text-003}}

\newcommand{\pplai}{\model{PerplexityAI}}

\newcommand{\checker}[1]{\textit{#1}\xspace}
\newcommand{\rarr}{\checker{RARR}}
\newcommand{\factscore}{\checker{FactScore}}
\newcommand{\factool}{\checker{FacTool}}
\newcommand{\factcheckgpt}{\checker{Factcheck-GPT}}

\newcommand{\longformsafe}{\checker{Longform SAFE}}

\newcommand{\fire}{\checker{FIRE}}

\newcommand{\dataset}[1]{\text{#1}\xspace}
\newcommand{\factqa}{\dataset{FactQA}}
\newcommand{\snowball}{\dataset{Snowball}}
\newcommand{\freshqa}{\dataset{FreshQA}}
\newcommand{\selfaware}{\dataset{SelfAware}}
\newcommand{\factoolqa}{\dataset{FacTool-QA}}
\newcommand{\felmwk}{\dataset{FELM-WK}}
\newcommand{\factcheckbench}{\dataset{Factcheck-Bench}}
\newcommand{\factscorebio}{\dataset{FactScore-Bio}}
\newcommand{\halueval}{\dataset{HaluEval}}
\newcommand{\factbench}{\dataset{FactBench}}

\newcommand{\module}[1]{\textsc{#1}\xspace}
\newcommand{\ofc}{\model{OpenFactCheck}}
\newcommand{\responseevaluator}{\module{ResponseEval}}
\newcommand{\llmevaluator}{\module{LLMEval}}
\newcommand{\factcheckerevaluator}{\module{CheckerEval}}

\usepackage{amssymb}
\usepackage{pifont}
\title{OpenFactCheck: A Unified Framework for Factuality Evaluation of LLMs}

\author{\textbf{Hasan Iqbal}\textsuperscript{1}\thanks{\xspace\xspace Equal contribution.} \quad  
        \textbf{Yuxia Wang}\textsuperscript{1}$^*$ \quad 
        \textbf{Minghan Wang}\textsuperscript{2}  \\ 
        \textbf{Georgi Georgiev}\textsuperscript{3} \quad 
        \textbf{Jiahui Geng}\textsuperscript{1} \quad
        \textbf{Iryna Gurevych}\textsuperscript{1} \quad
        \textbf{Preslav Nakov}\textsuperscript{1} \\
\textsuperscript{1}MBZUAI \quad 
\textsuperscript{2}Monash University \quad  
\textsuperscript{3}Sofia University \\
\texttt{\{hasan.iqbal, yuxia.wang, preslav.nakov\}@mbzuai.ac.ae}
}

\begin{document}
\maketitle
\begin{abstract}
The increased use of large language models (LLMs) across a variety of real-world applications calls for automatic tools to check the factual accuracy of their outputs, as LLMs often hallucinate. This is difficult as it requires assessing the factuality of free-form open-domain responses. While there has been a lot of research on this topic, different papers use different evaluation benchmarks and measures, which makes them hard to compare and hampers future progress. To mitigate these issues, we developed \textbf{\ofc}, a unified framework, with three modules:
(\emph{i})~\responseevaluator, which allows users to easily customize an automatic fact-checking system and to assess the factuality of all claims in an input document using that system,
(\emph{ii})~\llmevaluator, which assesses the overall factuality of an LLM, and (\emph{iii})~\factcheckerevaluator, a module to evaluate automatic fact-checking systems.
\ofc is open-sourced\footnote{\url{https://github.com/mbzuai-nlp/openfactcheck}} and publicly released as a Python library\footnote{\url{https://pypi.org/project/openfactcheck/}} and also as a web service\footnote{\url{http://app.openfactcheck.com}}. A video describing the system is available at \url{https://youtu.be/-i9VKL0HleI}.

\end{abstract}

\section{Introduction}




Large language models (LLMs) have demonstrated impressive capabilities in generating naturally-sounding answers over a broad range of human inquiries.
However, \gptfouro~\citep{gpt4} and other text generation models still produce content that deviates from real-world facts~\citep{bang2023multichatgpt, ali2023failurechatgpt, giuven2023llmsfailures}. 
This degrades the performance of LLMs and undermines their reliability, which is a significant bottleneck for their deployment~\citep{chuang2023dola, geng2023survey},
especially for high-stake applications, e.g., clinical, legal, and financial settings.

\begin{figure}[!t]
    \centering
    \includegraphics[width=1\linewidth]{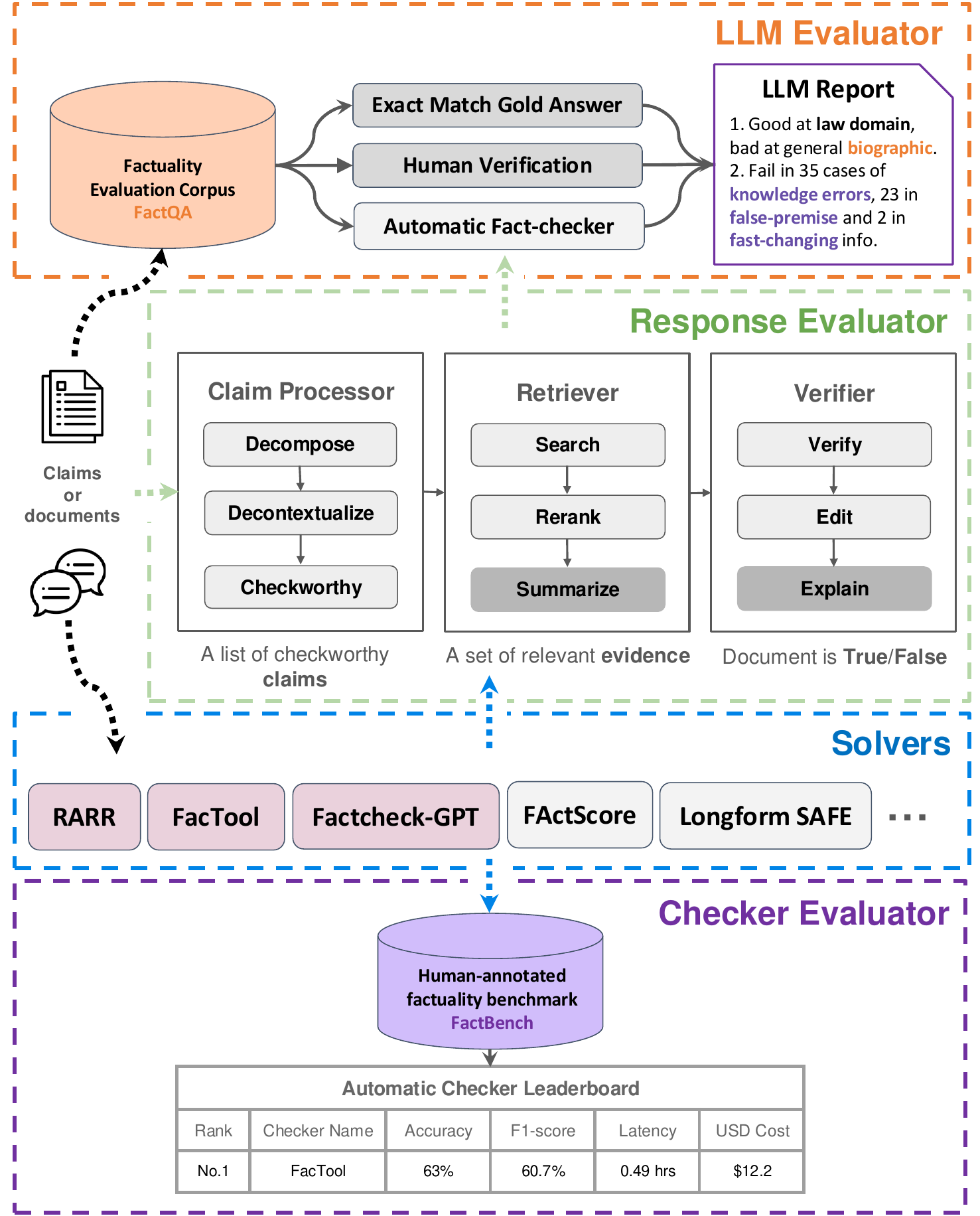}
    \caption{Overview of the \ofc demo system for LLM factuality evaluation and its modules. Green \responseevaluator: a customized fact-checker to identify factual errors given text inputs. Orange \llmevaluator: an LLM factuality evaluator to assess the LLM factual ability from different aspects and then to produce a report to illustrate its weaknesses and strengths. Purple \factcheckerevaluator: a fact-checker evaluator and leaderboard to encourage the development of advanced checkers in terms of performance, latency and costs.}
    \label{fig:framework}
\end{figure}

Many studies have explored evaluating the factuality of LLMs~\cite{lee2022factuality, chuang2023dola, shi2023trusting, chen2023felm}.
Two challenges have been identified: (\emph{i})~it is difficult to assess the factuality of open-domain free-form responses, and (\emph{ii})~different papers use different evaluation datasets and measures, which makes it hard to compare them, thus hampering future progress~\cite{wang2024factuality-survey}.
To mitigate these issues, we introduce \textbf{\ofc}.

\ofc is an Open-source Factuality Evaluation Framework for LLMs and it comprises the following three core modules (see \figref{fig:framework}):
\begin{itemize}
    \item \responseevaluator: It allows users to customize an automatic fact-checker and to verify the factuality of all claims made in a free-form document to alleviate the first problem. 
    \item \llmevaluator: A unified LLM factuality evaluation module which applies seven factuality-specific benchmarks to assess the LLM factuality ability from different aspects and then produces a report to illustrate the weakness and strength, tackling the second challenge.
    \item \factcheckerevaluator: It assesses the verification accuracy of fact-checkers, equipped with a leaderboard in terms of accuracy, latency, and costs, aiming to encourage the development of advanced automatic fact-checking systems.
\end{itemize}


The modules are designed for seamless integration, each contributing to and enhancing the capabilities of the others. The results of human verification derived from \llmevaluator can be used as the benchmark for evaluating the accuracy of automated fact-checkers. Simultaneously, the most effective checker identified in \factcheckerevaluator can be deployed for automated fact-checking tasks. Each fact-checker in \factcheckerevaluator can be an implementation in \responseevaluator. Complex user inquiries may be considered as potential candidates of the factuality assessment dataset utilized in \llmevaluator.

\textbf{General users} can tailor their checkers according to their specific needs, such as domain specialization, cost-effectiveness, or rapid processing, and identify factual errors for both human-written text (a claim or document) and the outputs of LLMs.\textbf{ LLM researchers and practitioners} can directly submit their LLM responses to the \llmevaluator by downloading our question set. Subsequently, we conduct evaluations to assess the model's factual accuracy and to generate a report analyzing the model performance from multiple aspects.
Similarly, \textbf{developers} who seek to evaluate and to fairly compare the efficacy of their fact-checking systems to other ones can upload their checker's verification outcomes to \factcheckerevaluator. Then, our system will show the ranking information in the leaderboard after evaluating under the same measurements.


To sum, three modules of \ofc respectively address the following:
\begin{itemize}
\item how to effectively identify factual errors in a text input;
\item how to systematically evaluate the factuality ability of an LLM;
\item which automatic fact-checker is the best, and which component dominates the final verification accuracy.
\end{itemize}

We have launched an open-source initiative that includes the development of a Python library and a web interface tailored to support three major functionalities. This foundation is expected to act as a catalyst for future advancements in factuality evaluation for LLMs.
We encourage extensive implementation of unique, effective, and robust claim processors, retrievers and verifiers within fact-checking pipelines, collections of challenging questions that LLMs tend to make factual errors, and human-annotated fine-grained verification examples. We believe that this will help to promote and to advance future research on LLM factuality.  

\section{Related Work}
\label{app:relatedwork}
While numerous automatic fact-checking systems have developed, such as \rarr, \factscore, \factool, \factcheckgpt,  \longformsafe and \fire~\cite{gao2022attributed, min2023factscore, chern2023factool, wang2023factcheck, wei2024longformsafe, xie2024fire}, they are often inaccessible to general users who lack a Python environment to compile code and run verification. Although these systems can function as the backend of a service, a user-friendly web interface is necessary to allow general users to verify text inputs by simply typing or copying text and clicking a \texttt{check} button. \ofc addresses this by providing an accessible web interface.

In addition, various fact-checking systems have distinct advantages. For instance, \factcheckgpt offers a fine-grained framework to involve all possible subtasks that could improve the fact-checking system, \factool uses a low-latency evidence retriever through asynchronous processing, and \factscore introduces a scoring metric that calculates the percentage of true claims in a given text, thereby quantitatively assessing the credibility of the input. \ofc integrates these components into a unified system~\cite{wang2024factuality-survey}.

Recent open-sourced demo system Loki~\citep{Loki} also aims to leverage strength of various automatic fact-checkers, while it emphasizes optimization a single fact-checking system in terms of accuracy, latency, robustness, cost-efficiency, and extensive support for multiple languages and LLMs.
In contrast, \ofc is a unified framework to cover three major functionalities for factuality evalaution of LLMs, including customizing a fact-checker by combining modules of different checkers, assessing LLM factuality from various perspectives, and evaluating the accuracy of automatic fact-checkers~\citep{wang2024openfactcheck}. 
\section{System Architecture}
\label{sec:system-architecture}
The design of \ofc emphasizes two principles: \emph{(i)} customizability and extensibility for both users and developers, and \emph{(ii)} compatibility with existing methods and datasets.
It consists of three modules: \responseevaluator, \llmevaluator, and \factcheckerevaluator. We detail the design and implementation of each components below.

\begin{table*}[t!]
  \centering
  \resizebox{\textwidth}{!}{
  \begin{tabular}{l|p{10cm}|p{6.5cm}|l r}
    \toprule
    \textbf{Dataset}$\downarrow$ & \textbf{The Ability to Evaluate} & \textbf{Domain} & \textbf{Error} & \textbf{Size} \\
    \midrule
    \textbf{\snowball} & Snowballing hallucination when model immediately output & Math, history, graph search & Type 2 & 1,500 \\
    \textbf{\selfaware} & Understand their own limitations on the unknowns & Biology, philosophy, psychology, history & Type 1,3 & 3,369 \\
    \textbf{\freshqa} & Answer questions changing fast over time or with false premises & Sports, entertainment, history, technology & Type 3 & 600 \\ 
    \textbf{\factoolqa} & Respond knowledge-based questions & History, geography, biology, science & Type 1 & 50 \\
    \textbf{\felmwk} & Answer world-knowledge questions & History, biology, geography, sports & Type 1 & 184 \\
    \textbf{\factcheckbench} & Answer open-domain, false-premise questions & Technology, history, science, sports & Type 1,2 & 94 \\
    \textbf{\factscorebio} & Generate detailed biographies & Biography & Type 1,3 & 683 \\
   \midrule
    \textbf{Total} & LLM factuality against world knowledge & 482 domains, top20 accounts for 70\% & Type 1,2,3 & 6,480 \\
    \bottomrule
  \end{tabular}
  }
  \caption{\textbf{\factqa}: factual vulnerability, domain, potential error type and size across seven component datasets.}
  \label{tab:dataset-statistics}
\end{table*}
\subsection{\responseevaluator}

\responseevaluator allows users to build a customized fact-checking system by selecting a claim processor, a retriever, and a verifier in web pages. The current version supports the following fact-checking systems: \rarr, \factool and \factcheckgpt~\citep{gao2022attributed, chern2023factool, wang2023factcheck}.

\paragraph{Configurable Architecture} 
We consolidate various fact-checking systems into a three-step process, encapsulated by three classes: \verb|claim_processor|, \verb|retriever|, and \verb|verifier|~\citep{wang2024factuality-survey}. These classes are instantiated and sequentially connected to form a pipeline that addresses the following tasks: \emph{(i)} breaking down a document into individual claims, \emph{(ii)} gathering pertinent evidence for each claim, and \emph{(iii)} evaluating the veracity of each claim based on the evidence provided. This sequence of tasks is referred to as \verb|solvers|. (see the pseudo code in \appref{app:pseudocode})

The implementation of a task solver can be flexible, just ensuring that the input and the output are aligned with the abstract class definitions. For example, evidence can be retrieved by calling SerpAPI or by searching Wikipedia using BM25, but we must return a list of relevant passages given an input claim.

Moreover, task solvers in our pipeline are not hard-coded, but can be configured through a \textit{YAML} configuration file. Thus, users can combine task-solver implementations from different systems (e.g., using \factcheckgpt's claim processor, \rarr's retriever, and \factool's verifier) and start the verification from any step. For example, users can start from the step of retrieval when the input does not need decomposition.

This functionality is achieved by a message-passing mechanism, where a \verb|success_flag| is used to indicate whether the current task solver successfully executes and returns the expected output.
The success flag passes through the pipeline as the configured order of solvers, guaranteeing that the output of the preceding solver fits the input for the current solver, otherwise error warning will be issued. 
Practically, the input and the output parameter names for the task solvers are defined in the configuration file. To link different solvers into a pipeline, one only needs to ensure that the current solver output name matches the input name of the succeeding solver. A \verb|FactcheckerState| class ensures storage of all information in the verification.

\paragraph{Extendable Architecture} Inspired by Fairseq, our framework is designed to be highly extendable by treating any third-party task solvers as plugins~\citep{ott2019fairseq}.
As long as the developed task solvers adhere to our class interface definitions, they can be imported and used in our framework.

\subsection{\llmevaluator}
\label{sec:llmevaluator}

We observed that studies assessing language models' factuality or evaluating whether the methods are effective to mitigate model hallucinations use different datasets and metrics.
This makes it difficult to compare, in the same conditions, the factuality of different models as well as to compare the effectiveness of different factuality enhancement approaches.
Moreover, a lot of prior work applied datasets such as MMLU~\citep{dan2021mmlu}, StrategyQA~\citep{geva2021strategyqa} and HotpotQA~\citep{yang2018hotpotqa} to evaluate model's factuality.
These datasets tend to focus on assessing the general performance, rather than factuality.
To this end, we first collect a dataset \textit{\factqa} by gathering factual questions of existing datasets that are curated to probe diverse factual errors and span across a spectrum of domains, to fairly evaluate LLMs' factuality under the same criteria

\paragraph{Factual Question Collection}
We collected factual questions from seven commonly-used corpora that is collected deliberately to assess LLM's factuality, including \snowball~\citep{zhang2023snowball}, \selfaware~\citep{yin-etal-2023-large}, \freshqa~\citep{vu2023freshllms}, \factool~\citep{chern2023factool}, \felmwk~\citep{chen2023felm}, \factcheckgpt~\citep{wang2023factcheck} and \factscorebio, a total of 6,480 examples shown in \tabref{tab:dataset-statistics}, referring to \factqa (see dataset details in \appref{app:factqadatasets}).

To concretely analyze models' vulnerability, we identify three labels for each question from the perspective of the knowledge domain, the topic, and the potential error type if a LLM generates a factually incorrect response. 
So each example includes the following fields: \textit{question}, \textit{domain}, \textit{topic}, \textit{ability to test}, \textit{task} and \textit{source}. 
Domains involve general, legal, biomedical, clinical, scientific and so on. Given a domain, we further fine-grained topics. 
Three common error types are presented.

\textbf{\ftype: \textit{Knowledge error}} is the most common error when the model produces hallucinated or inaccurate information due to lacking relevant knowledge or internalizing false knowledge in the pre-training stage or in the alignment process.

\textbf{\stype: \textit{Over-commitment error}} occurs when the model fails to recognize the falsehoods (or jokes) in the prompt or previously-generated context, and provides an inaccurate or inappropriate response.

\textbf{\ttype: \textit{Disability error}} happens when the model is unable to search up-to-date information to correctly answer questions whose answers change over time, e.g., \textit{What is today's gas price in New York} (fast-changing). See more in \appref{app:errortype}.

\paragraph{Evaluation Measurement}
For questions that can be answered by Yes/No or have a short gold answer, we perform exact matching between the model responses and the gold standard answer to judge whether the response is factually correct, and then to calculate accuracy, such as for \snowball and \selfaware.

For \freshqa, we use the \textit{FreshEval} proposed in \citet{vu2023freshllms} to evaluate the correctness of model's responses.
For open-domain questions from the other four datasets with free-form and long responses, there are no gold standard answers. We use automatic fact-checking systems to judge the correctness of claims and obtain the percentage of true claims as the accuracy for a response.

\subsection{\factcheckerevaluator}
\label{sec:factcheckerevaluator}
Automatic fact-checking systems aim to identify whether a claim or a document is true or false, but the results are not necessarily correct.
To assess the accuracy of automatic fact-checkers, we gather four LLM factuality benchmarks with human-annotated factual labels for three levels of granularity text: claims/segments/documents given (question, \chatgpt response) pairs, including \factoolqa, \felmwk, \factcheckbench and \halueval as shown in \tabref{tab:factbench-statistics}. 
We refer to them as \factbench.
We use precision, recall, and F1-score with respect to the \textit{True} or \textit{False} claim/document to evaluate the effectiveness of fact-checking systems.



\begin{table}[t!]
\centering
\resizebox{0.94\columnwidth}{!}{
\begin{tabular}{l|rrr|r}
\toprule
\textbf{Dataset} $\downarrow$  & \textbf{\#True} & \textbf{\#False} & \textbf{\#Unknown} & \textbf{Total} \\
\midrule
\textbf{\factoolqa} & 177 & 56 & 0 & 233 \\
\textbf{\felmwk} & 385 & 147 & 0 & 532  \\
\textbf{\factcheckbench} & 472 & 159 & 47 & 678 \\
\midrule
\textbf{\halueval }& 3,692 & 815 & 0 & 4,507 \\
\bottomrule
\end{tabular}
}
\caption{The number of true, false claims and unknown (no-enough-evidence or opinions) for \factoolqa, \felmwk and \factcheckbench, the number of responses for \halueval (no claim-level labels).}
\label{tab:factbench-statistics}
\end{table}

\paragraph{Discussion about \textit{Unifying}}
It can be argued that the underlying philosophies of the three modules differ, reflecting varying interpretations of factuality. For example, the design view o \llmevaluator and \factcheckerevaluator differs from that of \responseevaluator. 

Our goal is to integrate all LLM factuality evaluation functionality into a unified framework, while preserving the individual function.

The \llmevaluator employs different metrics across datasets. This may be debated. Similarly, we aim to consolidate these datasets into a unified benchmark, enabling other studies to utilize a standardized evaluation function. This approach would enhance the fairness of comparisons across studies, as they would be evaluated consistently, despite the use of dataset-specific evaluation measures.

We acknowledge the limitation of the current \factcheckerevaluator, which is restricted to evaluating the verification step. We plan to progressively extend its capabilities to support fine-grained evaluations across multiple steps.\footnote{The evaluator currently supports both claim-level and document-level verification, depending on whether users download claim or document datasets.}

\section{Access and Deployment}
\ofc is accessible via a user-friendly web interface and features an integrated database that maintains a user leaderboard. It is also available as a standalone open-source Python library. 

\subsection{Python Library}
\ofc is available as an open-source Python library on PyPI, designed for flexibility and ease of integration into existing projects. This library equips developers with essential components for fact-checking in any Python environment, making it an optimal choice for enhancing applications with fact-checking features. The library employs a fluent interface to ensure its usage is intuitive for both beginners and experts alike.

Users can install the library by simply using the pip package manager:
\begin{minted}{bash}
$ pip install openfactcheck
\end{minted}

The library includes detailed documentation to assist developers in customizing and extending the functionality to meet their specific needs and it is continually updated to ensure compatibility with the latest research and data security standards. 

\paragraph{Usage Examples}
The first step is to import the necessary library components and initialize \verb|OpenFactCheckConfig| configuration and \verb|OpenFactCheck| class, which requires no input values for default usage, as shown below:
\begin{minted}{python}
from openfactcheck import OpenFactCheck, OpenFactCheckConfig
config = OpenFactCheckConfig()
ofc = OpenFactCheck(config)
\end{minted}

Upon importing the library, users are required to secure API keys from platforms utilized by OpenFactCheck’s default solvers for evidence retrieval and claim verification. These keys are available from OpenAI\footnote{\url{https://openai.com/api}}, SerpAPI\footnote{\url{https://serpapi.com}}, and ScraperAPI\footnote{\url{https://scraperapi.com}}. After acquiring the keys, they need to be configured as environment variables to enable their use within the library.

The three key functionalities outlined in \secref{sec:system-architecture} are implemented as shown in \figref{fig:usage}.
We can see that the design of the library is intuitive and straightforward, enabling users to apply it without extensive learning, and practioners to perform further developments easily (e.g., reusing one example by simply altering the evaluator name in each instance). 
The intermediate results are also logged on the terminal and are omitted here for brevity.

User is provided with the benchmarks for the LLM and FactChecker evaluations, and can upload the responses to the library for evaluation in the form of CSV files. CSV file format for LLM evaluation has two columns: \texttt{index} and \texttt{response}, while the FactChecker evaluation CSV file format has three columns: \texttt{label}, \texttt{time}, and \texttt{cost}.

\begin{figure}[t!]
\centering
\begin{minted}{python}
ofc.ResponseEvaluator.evaluate(response: str)
# response: string output from LLM
\end{minted}

\begin{minted}{python}
ofc.LLMEvaluator.evaluate(model_name: str, input_path: str)
# model_name: evaluated model name.
# input_path: path to the CSV containing responses for the LLM Benchmark.

# Output
# A dictionary with detailed scores (precision, recall, f1, accuracy, cost, time etc. for each dataset subset i.e. snowballing, selfaware, freshqa, factoolqa, felm-wk, factcheck-bench and factscore-bio.
\end{minted}

\begin{minted}{python}
ofc.CheckerEvaluator.evaluate(input_path: str)
# input_path: path to the CSV containing responses for the FactChecker Benchmark

# Output
# A dictionary with detailed scores (precision, recall, f1, accuracy, cost, time etc.)
\end{minted}
    \caption{Usage examples of three major modules: \responseevaluator, \llmevaluator and \factcheckerevaluator.}
    \label{fig:usage}
\end{figure}

\begin{figure*}[t!]
    \centering
    \includegraphics[height=0.9\textheight]{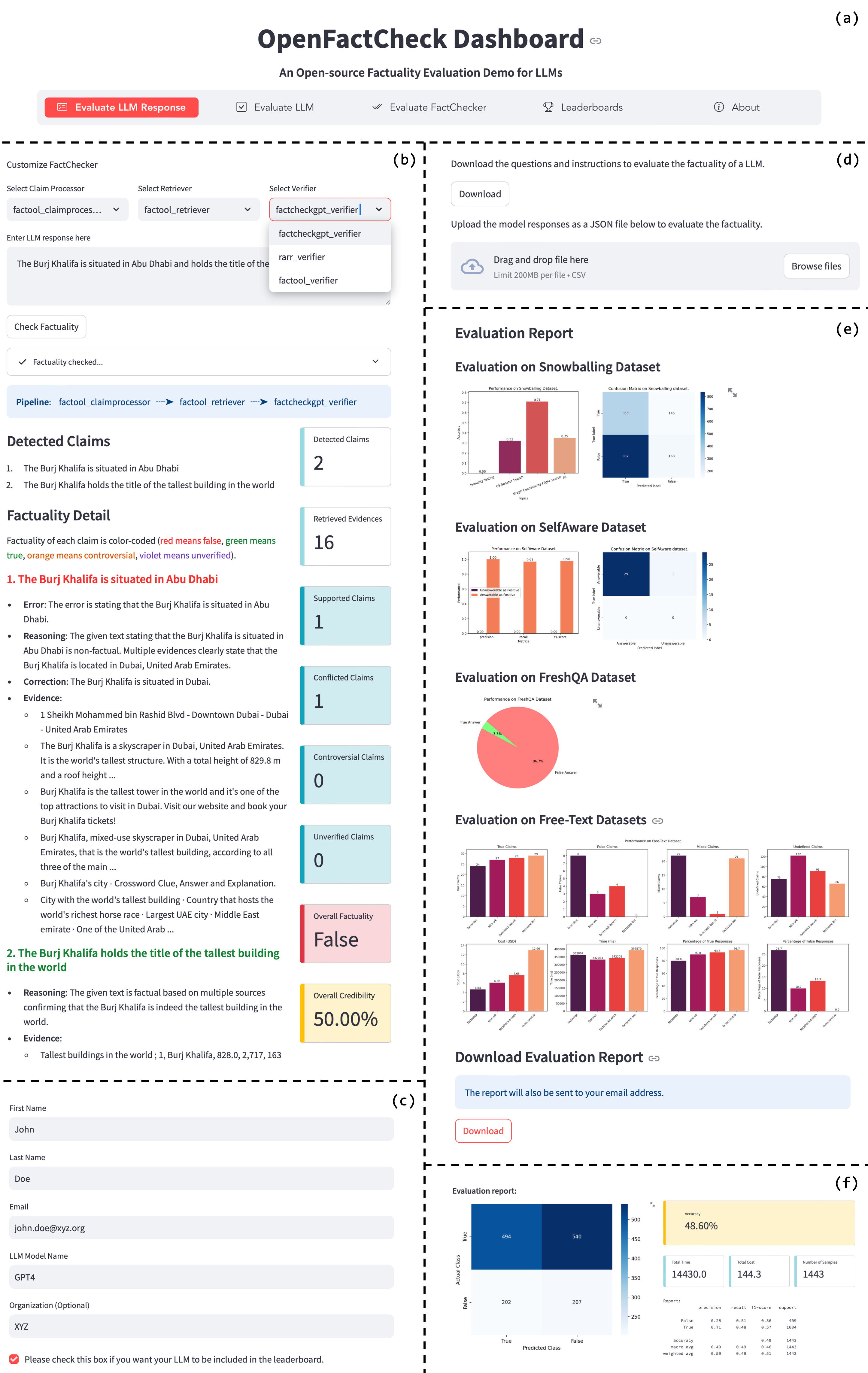}
    \caption{\ofc Dashboard: \textbf{(a)} is the navigation bar. \textbf{(b)} a claim processor breaking down the input into two atomic claims. The retriever collected 16 pieces of evidence, and the verifier assessed each claim individually, with one true and one false, resulting 50\% credibility overall. \textbf{(c)} shows the user information required before uploading LLM responses or verification results to \llmevaluator and \factcheckerevaluator. \textbf{(d)} shows the functions of downloading and uploading. \textbf{(e)} and \textbf{(f)} exhibit the LLM and FactChecker Evaluation report respectively.}
    \label{fig:dashboard}
\end{figure*}

\subsection{Web Interface}
The web interface of \ofc provides a user-friendly platform that allows general users to interactively engage with the fact-checking functionalities. It is designed to accommodate both novice and expert users, facilitating easy access to the comprehensive evaluations involved in the assessment of LLM factuality. The web interfaces are organized into four distinct sections as illustrated in \figref{fig:dashboard}(a). 

In \responseevaluator page as shown in \figref{fig:dashboard} (b), users can click the drop\-down list to select from a range of pre-implemented \texttt{claim processor}, \texttt{retriever}, and \texttt{verifier}. Then, users can input text either written by human or generated by machine into the text box and click \textit{Check Factuality} to obtain the verification results. 
As the example demonstrated in the Figure, it includes two claims. 

The system collected sixteen pieces of evidence, and one claims is supported and one claim is refuted, resulting the overall credibility of 50\% and judgement ``False'' for this whole input.

For both the \llmevaluator and \responseevaluator pages exhibited in \figref{fig:dashboard} (d), users first download either the question set \factqa or the claims/documents in \factbench. After being ready to upload the responses of the LLM that users aim to assess or the verification results of the fact-checkers to test, users type their details including name, email address and so on, and provide the option to opt in or out of leaderboard inclusion (see \figref{fig:dashboard} (d)). If users agree, their information and rank will be displayed on the leaderboard, otherwise invisible for others. 

It may takes some time for \llmevaluator to generate teh evaluation report, depending on the system’s current load. Once the report is ready, it is emailed directly to the user, eliminating the need to wait within the application.
LLM factuality evaluation report presents LLM factuality from various aspects, and specifically includes accuracy and confusion matrix of short answers, pie chart indicating accuracy over fresh questions and bar chart showing the percentage of true, false, controversial claims for free-form responses, as shown on \figref{fig:dashboard} (e). 

Similarly, \factcheckerevaluator results present the number of evaluated examples, the overall accuracy, total time and USD cost, fine-grained precision, recall and F1-score for false and true classes, and a confusion matrix showing the mis-identification of this fact-checker. The submission in \figref{fig:dashboard} (f) reveals that this checker performs equally poor over both false and true claims in verification.
This evaluation is instant. \footnote{See more evaluation results in \citet{wang2024openfactcheck}.}

\section{Conclusion and Future Work}

We implemented a unified, easy-to-use and extensible framework \ofc. It is accessible by both Python library and web service, supporting the customization and evaluation of automatic fact-checking systems and LLM factuality evaluation.
Specifically, \ofc allows general users to check whether a claim and a document are factual or not by clicking \textbf{Check}, and also facilitate LLM practitioners and developers to effectively and efficiently evaluate the factuality of their LLMs from various perspectives, and to assess the accuracy of automatic fact-checking systems. 
In the future, we will continue to integrate new techniques, features, and evaluation benchmarks to \ofc to facilitate the research progress of LLM fact-checking.

\section*{Limitations and Future Work}
While \ofc presents a comprehensive framework for factuality evaluation of LLMs, several limitations must be acknowledged:

\paragraph{Multilingual Expansion}
\ofc is a platform that combines the features of various fact-checking systems and is designed to be language-agnostic. While the default task solvers in the system are configured for English, the platform can be expanded to accommodate other languages by developing task solvers that align with the specific linguistic requirements of those languages. This flexibility allows for easy adaptation and extension to support multilingual fact-checking capabilities.

\paragraph{Evaluation Datasets}
The effectiveness of \ofc is dependent on the quality and diversity of the datasets used for evaluation. While we have integrated multiple datasets to cover a broad spectrum of domains and potential factual errors, the evaluation is still limited by the inherent biases and coverage gaps in these datasets. For instance, some specialized domains may not be adequately represented, potentially affecting the robustness of the evaluation for LLMs in those areas.

\paragraph{Latency and Costs}
The performance of automatic fact-checking systems integrated within \ofc can vary significantly in terms of latency and operational costs. High accuracy often comes at the expense of increased computational resources and processing time, which may not be feasible for all users, particularly those with limited budgets or time constraints.

\paragraph{Reliance on External Knowledge Sources}
The fact-checking modules depend heavily on external knowledge sources, such as Wikipedia and web search engines. The availability and reliability of these sources can affect the accuracy and completeness of the fact-checking process. Furthermore, the dynamic nature of web content means that the information retrieved may not always be up-to-date.

\paragraph{Temporal Issues}
The factuality of statements can change over time due to evolving events, new discoveries, or updated information. \ofc does not explicitly account for temporal dynamics as of now, which may lead to discrepancies between the evaluation results and the current state of knowledge. Authors are already working on factuality evaluation methods that consider temporal aspects, which will be integrated into \ofc in future releases.

\section*{Ethical Statement}
The development and deployment of \ofc are guided by a commitment to ethical principles, ensuring that the framework is used responsibly and for the benefit of society:

\paragraph{Transparency and Accountability}
We strive to maintain transparency in the design, implementation, and evaluation of \ofc. The source code and datasets are publicly available, enabling scrutiny and fostering trust within the research community. We encourage users to report any issues or biases they encounter, facilitating continuous improvement.

\paragraph{Bias Mitigation}
Recognizing that biases can exist in both datasets and LLMs, we are dedicated to minimizing such biases in \ofc. By integrating diverse evaluation benchmarks and encouraging the development of fair fact-checking approaches, we aim to reduce the impact of biases on factuality evaluation outcomes.

\paragraph{Social Impact}
By enhancing the factual accuracy of LLMs, \ofc aims to contribute positively to society. Accurate information is crucial for informed decision-making and public discourse. We believe that improving the reliability of LLM outputs can help combat misinformation and support the dissemination of truthful information.

\bibliography{ref}
\bibliographystyle{acl_natbib}
\clearpage
\appendix

\section{Pseudo Code of \responseevaluator}
\label{app:pseudocode}

In this section, we present the pseudo code for the \responseevaluator, a modular system designed to process, retrieve, and verify claims found in textual documents. The system is divided into three primary components: the claim processor, the retriever, and the verifier. Each module is tasked with a specific function—extracting claims from the input document, retrieving relevant evidence, and verifying the factual accuracy of the claims, respectively. Figure \ref{fig:python_code} outlines the pseudo code implementation of each module, showcasing the flow from document processing to final verification. This structured approach allows for a systematic handling of claims, leveraging both natural language processing tools and deep learning models to ensure a comprehensive evaluation of document veracity.

\begin{figure}[ht!]
\centering
\begin{minted}{python}
def claim_processor(document: str) -> List[str]:
    # FactScore
    paragraphs = documents.split("\n")
    sentences = [NLTK(para) for para in paragraphs]
    claims = [call_LLM(sentence, prompt="decompose into atomic claims") for sentence in sentences]

    # FacTool
    claims = call_LLM(document, promot="extract context-independent atomic claims based on the document")
    
    return claims
    
def retriever(claim: str, database: DB, retrieval_strategy: obj, search_api_key: str) -> List[str]:
    # offline DB dump
    evidence = retrieval_strategy(claim, database)
    
    # online web pages by calling API
    evidence = serper_or_serpapi(claim, search_api_key)

    return evidence

def verifier(claim: str, evidence: List[str]) -> bool:
    # call LLMs
    factual_label = call_LLM(claim, evidence, prompt="based on the evidence and your own knowledge, determine whether the claim is true or false.")

    # use NLI models
    stance2factual = {
        "entailment": true,
        "contradiction": false,
        "neutral": "not enough evidence"
    }
    stances = [nli(evid, claim) for evid in evidence]
    majority_stance = majority_vote(factual_labels)
    factual_label = stance2factual[majority_stance]
    
    return factual_label
\end{minted}
    \caption{Pseudo code for classes in \responseevaluator.}
    \label{fig:python_code}
\end{figure}

\section{Factual Error Evaluation}
\label{app:errortype}
\textbf{\ftype: \textit{Knowledge error}} is the most common error, occurring when the model produces hallucinated or inaccurate information.
However, LLMs do not know what they do not know, sometimes overestimate their capacities and confidently output unknown information, leading to false responses.
Mitigating such errors require: (a) learning and correcting parametric knowledge through the curation of corpora used in pre-training, supervised fine-tuning (SFT) and alignment, (b) augmenting by external knowledge in inference, (c) calibrating models to be aware of unknowns, and (d) configuring the decoding strategies (sample/beam-search, temperature), balancing diversity and accuracy~\cite{zhang2023siren}.

\textbf{\stype: \textit{Over-commitment error}} occurs when the model fails to recognize the falsehoods (or jokes) inherent in the prompt or previously-generated context, and provides an inaccurate or inappropriate response.
The left-to-right generation strategy used by LLMs poses potential risks that LLMs sometimes over-commit to the false premise in the context, even when they recognize they are incorrect~\citep{zhang2023siren}.
To address this issue, engineering better prompts is helpful, such as explicitly instructing models to first detect false premises in the prompt~\citep{vu2023freshllms} and asking the same question in a different way (\textit{Is 10733 a prime number?} $\rightarrow$ \textit{What are the factors of 10733? Let's think step-by-step.}) 
\newpage

\textbf{\ttype: \textit{Disability error}} happens when the model is unable to search up-to-date information to correctly answer questions whose answers change over time, e.g., \textit{What is today's gas price in New York} (fast-changing).
Retrieving external knowledge and augmenting it in the context would help for such cases. 
Note that we do not consider \textit{reasoning errors} that arise when a claim is based on flawed reasoning or faulty logic.

Thus, ex exclude \textit{irrelevant error} concerning that the content is unrelated to the question~\citep{chen2023felm}. The former highlights LLM's reasoning ability, which is more reflected in math and reasoning tasks, and the latter has more to do with response's helpfulness or human preference. 
They are important in LLM evaluation, and may implicitly influence factuality, but we will first focus on explicit causes, leaving the implicit for future work. 

\begin{table}[t!]
  \centering
  \resizebox{\columnwidth}{!}{
  \begin{tabular}{lr|lr}
    \toprule
    \textbf{Domain} & \textbf{Size} & \textbf{Domain} & \textbf{Size}\\
    \midrule 
    History          & 771 & Science          & 143 \\
    Biography        & 683 & Physics          & 136\\
    Mathematics      & 612 & Social Sciences   & 111\\
    Transportation   & 519 & Literature       & 100\\
    Biology          & 259 & Geography        & 87\\
    Philosophy       & 229 & Astronomy        & 82\\
    Technology       & 208 & Economics        & 69\\
    Entertainment    & 191 & Music            & 66 \\
    Psychology       & 169 & Religion         & 63\\
    Sports           & 157 & General Knowledge & 53\\
    \midrule 
    \textbf{Total} & & \multicolumn{2}{r}{\textbf{4,523 (69.8\%)}}\\
    \bottomrule
  \end{tabular}}
  \caption{\factqa's top-20 domains and the number of examples from each domain.}
  \label{tab:domain-dist}
\end{table}

\section{\factqa Component Datasets}
\label{app:factqadatasets}
\textbf{\snowball} dataset~\cite{zhang2023snowball} comprises three question-answering subsets: primality testing, senator search, and graph connectivity, each with 500 yes/no questions. They aim to investigate snowballing hallucination when a model immediately outputs an incorrect answer (yes or no) as false generated context. Language models are prompted to first output a yes/no answer and then to provide explanations. When the immediate answer is wrong, the model tends to continue to snowball the false statements instead of correcting them.
%

\textbf{\selfaware}~\citep{yin-etal-2023-large} aims to evaluate LLMs' ability to understand their own limitations and unknowns. This is achieved by assessing models' ability to identify unanswerable or unknowable questions. They compiled a collection of 1,032 unanswerable questions from online platforms like Quora and HowStuffWorks. In addition, they gathered 2,337 answerable questions from sources such as SQuAD, HotpotQA, and TriviaQA, resulting in a total of 3,369 questions.

\textbf{\freshqa}~\cite{vu2023freshllms} is composed of 600 natural, open-ended questions, segmented into four primary categories based on the answer's stability: \emph{never-changing}, for answers that rarely alter, \emph{slow-changing}, for those that evolve over several years, \emph{fast-changing}, for answers that shift within a year or less, and \emph{false-premise}, encompassing questions with factually incorrect premises that need to be countered.

\textbf{FacTool}~\cite{chern2023factool} detected factual errors in LLM generations across four different tasks: knowledge-based QA, code generation, mathematical reasoning, and scientific literature review. 
We used 50 knowledge-based QA \factoolqa in \factqa.
 
\textbf{FELM}~\cite{chen2023felm} 
collects responses generated from LLMs and annotated factuality labels in a fine-grained manner. The dataset consists of 5 categories, with examples per category as follows: 194 math, 208 reasoning, 125 science, 184 world knowledge (wk), and 136 writing recordings. We used 184 world-knowledge questions, referring to \felmwk.

\textbf{\factcheckbench}~\citep{wang2023factcheck} Factcheck-GPT gathered a total of 94 highly challenging questions from sources including Twitter posts, internal brainstorming, and Dolly-15k, encompassing 678 claims.

\textbf{\factscorebio}~\cite{min2023factscore} selected 183 entities, and collected responses from three LLMs including \davinci, \chatgpt, and \pplai, and then annotated factual labels (supported, not-supported and irrelevant) for each atomic claim by humans.

\end{document}